\documentclass{article} %

\PassOptionsToPackage{svgnames,table,dvipsnames}{xcolor}

\usepackage[final]{colm2025_conference}

\usepackage{microtype}
\usepackage{hyperref}
\usepackage{url}
\usepackage{booktabs}

\usepackage{lineno}
\usepackage[svgnames,table,dvipsnames]{xcolor}
\usepackage{microtype}
\usepackage{graphicx}
\usepackage{booktabs} %

\usepackage{multirow}
\usepackage{multicol}
\usepackage{enumitem}
\usepackage{algorithm}
\usepackage{algorithmic}
\usepackage{subcaption}
\usepackage{comment}
\usepackage{array}
\usepackage{booktabs}
\usepackage{xspace}
\usepackage{arydshln}
\usepackage[most]{tcolorbox}
\usepackage{amssymb}

\usepackage{url}

\usepackage{breakurl}
\usepackage{hyperref}

\usepackage{ifthen}

\usepackage{hyperref}

\usepackage{tabularx}
\usepackage{makecell} %
\usepackage{cleveref}

\newboolean{publicversion}
\setboolean{publicversion}{false}

\newcommand{\myx}{$\times$\xspace}

\newcommand{\arxivsum}{Arxiv Summarization\xspace}

\newcommand{\vheading}[1]{\vspace{0.05in}\noindent\textbf{#1.}}

\newcommand{\begincompactitemize}{\begin{itemize}[noitemsep,topsep=0pt,parsep=0pt,partopsep=0pt]}

\ifthenelse{\boolean{publicversion}}{
	\newcommand{\grumbler}[3]{}
        \newcommand{\jm}[1]{}
        \newcommand{\ap}[1]{}
        \newcommand{\nk}[1]{}
        \newcommand{\rr}[1]{}
        \newcommand{\sk}[1]{}
        \newcommand{\nitin}[1]{}
        \newcommand{\anmol}[1]{}
        \newcommand{\amey}[1]{}
        \newcommand{\alexey}[1]{}
        \newcommand{\todo}[1]{}
}
{
\newcommand{\grumbler}[3]{\xspace\textcolor{#3}{\bf #1: #2}}
\newcommand{\jm}[1]{\grumbler{Jayashree}{#1}{magenta}}
\newcommand{\ap}[1]{\grumbler{Ashish}{#1}{violet}}
\newcommand{\nk}[1]{\grumbler{Nipun}{#1}{purple}}
\newcommand{\sk}[1]{\grumbler{sk}{#1}{purple}}
\newcommand{\alexey}[1]{\grumbler{AT}{#1}{green}}
\newcommand{\nitin}[1]{\grumbler{Nitin}{#1}{orange}}
\newcommand{\rr}[1]{\grumbler{Ram}{#1}{cyan}}
\newcommand{\anmol}[1]{\grumbler{Anmol}{#1}{cyan}}
\newcommand{\amey}[1]{\grumbler{Amey}{#1}{teal}}
\newcommand{\todo}[1]{\textcolor{red}{#1}}
}

\newtcolorbox{checklistitem}{
    colback=gray!10,
    colframe=black!80,
    boxrule=0.8pt,
    title={\small Evaluation Checklist},
    fonttitle=\bfseries,
    left=6pt,
    top=1pt,
    bottom=1pt,
    before upper = {\greencheck\xspace}
}

\definecolor{codegreen}{rgb}{0,0.6,0}

\newcommand{\greencheck}{\textcolor{codegreen}{\checkmark}}

\newcommand{\redcross}{\textcolor{red}{$\times$}}

\definecolor{darkblue}{rgb}{0, 0, 0.5}
\hypersetup{colorlinks=true, citecolor=darkblue, linkcolor=darkblue, urlcolor=darkblue}

\title{On Evaluating Performance of LLM Inference Systems}

\author{Amey Agrawal$^{\text{1}}$\enskip Nitin Kedia$^{\text{2}}$\enskip Anmol Agarwal$^{\text{1}}$\enskip Jayashree Mohan$^{\text{2}}$\enskip \\
\textbf{Nipun Kwatra$^{\text{2}}$\enskip Souvik Kundu$^{\text{3}}$\enskip Ramchandran Ramjee$^{\text{2}}$\enskip Alexey Tumanov$^{\text{1}}$} \\ \\
\vspace{1mm} $^{\text{1}}$Georgia Institute of Technology \enskip $^{\text{2}}$Microsoft Research \enskip $^{\text{3}}$Intel Labs}

\begin{document}

\maketitle

\begin{abstract}
The rapid evolution of Large Language Model (LLM) inference systems has yielded significant performance and efficiency improvements. However, our systematic analysis reveals that current evaluation methodologies frequently exhibit fundamental flaws, often manifesting as common evaluation anti-patterns that obscure true performance characteristics and impede scientific progress. Through a comprehensive examination of recent systems, we identify recurring anti-patterns across three key dimensions: \textit{Baseline Fairness, Evaluation Setup, and Metric Design}. These anti-patterns are uniquely problematic for LLM inference due to its dual-phase nature combining distinct prefill and decode operations, its handling of highly heterogeneous workloads, and its strict temporal requirements for interactive use. We demonstrate how common anti-patterns—such as inadequate baseline comparisons that conflate engineering effort with algorithmic novelty, workload selections that fail to represent production scenarios, and metric normalizations that hide substantial performance variability like generation stalls—lead to misleading conclusions. To address these challenges, we provide a \textit{comprehensive checklist} derived from our analysis, establishing a framework for recognizing and avoiding these anti-patterns in favor of robust LLM inference evaluation. To demonstrate the practical application of our framework, we present a \textit{case study} analyzing speculative decoding, a technique whose bursty, non-uniform token generation is easily misinterpreted when evaluated using approaches characteristic of these anti-patterns. Our work establishes a rigorous foundation for evaluation methodology, enabling meaningful comparisons, ensuring reproducible results, and ultimately accelerating genuine progress in LLM inference systems by moving beyond common anti-patterns to align evaluation with real-world requirements.
\end{abstract}

\section{Introduction}

The past two years have witnessed remarkable progress in Large Language Model (LLM) inference systems. The community has dramatically improved efficiency, achieving orders-of-magnitude higher throughput and lower latency through innovations like continuous batching, paged attention, and speculative decoding \cite{orca,vllmsosp,agrawal2024taming,distserve2024,patel2023splitwise}. Open-source frameworks and commercial services have scaled these systems to millions, democratizing access to powerful generative AI.

This rapid technological progress, however, has outpaced our evaluation methodologies. Our systematic analysis of recent systems reveals a concerning \textit{divergence}: while the systems themselves have evolved rapidly, operating at unprecedented scales, our methods for evaluating them remain largely static. This stagnation leads to inconsistencies that obscure genuine performance characteristics and impede scientific advancement. Through a comprehensive examination of influential works from top-tier peer-reviewed venues, we identify some consistent anti-patterns in the evaluation of LLM inference systems.

These anti-patterns arise directly from the inherent complexities and unique characteristics of LLM inference serving. Specifically, LLMs combine a \textit{compute-intensive parallel prefill phase} for the prompt with a \textit{memory-bandwidth-bound sequential decode phase} for token generation, creating complex resource contention and performance dynamics (\textit{Dual-Phase Inference}). Furthermore, production deployments face extreme \textit{Workload Heterogeneity}, with vast diversity in prompt lengths (spanning tens to tens of thousands of tokens), output lengths, and varying strictness of latency needs depending on the application (e.g., interactive chat vs. batch summarization), which makes appropriate workload selection extremely critical. User experience often hinges critically on \textit{Latency Requirements}, such as time-to-first-token (TTFT) and consistent time-between-tokens (TBT), aspects easily obscured by simple end-to-end latency figures. Compounding these issues is the \textit{Rapid Architectural Evolution} of models (e.g., Grouped Query Attention (GQA) by \cite{ainslie-etal-2023-gqa}, Mixture-of-Experts (MoE) by \cite{shazeer2017outrageously}), which continually shift performance bottlenecks and demand adaptable evaluation strategies that move beyond outdated assumptions.

To improve LLM inference systems, we must first improve how we measure them. Our work provides a systematic framework for robust evaluation practices, centered on a comprehensive checklist that guides researchers through critical considerations such as workload selection and metric design. This checklist, derived from our analysis of common pitfalls, offers concrete guidance for ensuring that experimental evaluations accurately capture system performance in ways that correspond to real-world requirements. To illustrate the application of our methodological framework, we examine speculative decoding --- a popular decode latency optimization where naive evaluation approaches fail. This examination serves as a practical demonstration of how our evaluation principles can illuminate performance characteristics that otherwise remain hidden behind misleading statistics.

We identify consistent evaluation anti-patterns across three dimensions in the literature -- \textit{Baseline Fairness}, \textit{Evaluation Setup}, and \textit{Metric Design} -- that stem directly from these challenges. These recurring suboptimal practices frequently mask critical performance limitations, such as high scheduling delays or generation stalls, and can inflate the perceived benefits of new techniques. These anti-patterns manifest due to both the neglect of fundamental evaluation principles, which become particularly acute in LLM systems, and a failure to adapt methodologies to the unique characteristics of such systems outlined above.

This work aims to establish rigorous evaluation standards by:

\begin{itemize}[nolistsep]
\item Performing a systematic analysis of common evaluation anti-patterns in LLM inference, examining how these widespread practices lead to incorrect conclusions and identifying \textit{why} they are particularly problematic in the LLM context.
\item Providing a comprehensive evaluation checklist offering concrete guidance for robust evaluation, effectively serving as a tool to detect and avoid these anti-patterns, covering workload selection, latency bounds, and metric interpretation.
\item Presenting an empirical analysis using speculative decoding as a case study, demonstrating how avoiding evaluation anti-patterns reveals critical performance trade-offs obscured by conventional, yet flawed, approaches.
\end{itemize}

\section{Background}
\label{sec:background}

In this section, we describe the typical LLM inference process, commonly used metrics to characterize inference performance, and an overview of unique properties of LLMs inference systems that makes their evaluation challenging.

\subsection{LLM Inference Process}
\label{sec:background:inference-process}

\begin{figure}[t!]
\centering
\includegraphics[width=0.55\linewidth]{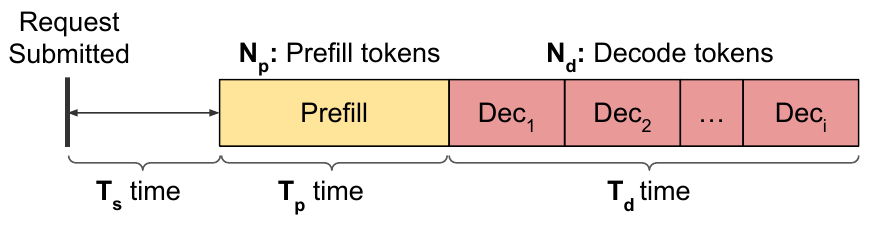}
\caption{Anatomy of an LLM inference request showing the three distinct phases: scheduling ($T_s$), prefill processing ($T_p$), and decode generation ($T_d$). The prefill phase processes $N_p$ input tokens in parallel, while the decode phase generates $N_d$ output tokens sequentially. Each decode token (Dec$_1$, Dec$_2$, ..., Dec$_i$) is generated one at a time.}
\label{sec:back:latbreakdown}
\end{figure}

There are two distinct phases in LLM inference -- a prefill phase followed by a decode phase. During prefill phase, the user's input prompt is processed and first output token is produced. Next, during the decode phase, output tokens are generated one at a time, where the token generated in one step is passed through the model to generate a new token in the next step until a special end-of-sequence token is generated. \autoref{sec:back:latbreakdown} illustrates this autoregressive generation process. Decode phase also requires access to KV (key and value) pairs associated with all previously processed tokens during its attention phase. %

\subsection{Performance Metrics for LLM Inference}
\label{sec:background:metrics}

Conventional performance metrics for LLM inference performance are the following:

\begin{itemize}[noitemsep,topsep=0pt,parsep=0pt,partopsep=0pt,leftmargin=*]
\item\textbf{Time To First Token (TTFT) \citep{distserve2024, llmperfdatabricks}}: Latency between request arrival and first output token, including scheduling delay ($T_s$) and prompt processing time. Critical for interactive applications requiring responsiveness.

\item\textbf{Time to Last Token (TTLT) \citep{orca,vllmsosp}}: End-to-end latency ($T_s + T_p + T_d$) for complete request processing. Essential for applications like code completion where partial outputs have limited utility.

\item\textbf{Time Between Tokens (TBT) \citep{agrawal2024taming}}: Latency of each subsequent token generation ($T^i_d$) during decoding. Directly impacts perceived model speed, with ~6 tokens/second matching typical reading speed.

\item\textbf{Time Per Output Token (TPOT) \citep{distserve2024, llmperfdatabricks}}: Average decode time per token ($T_d/N_d$).

\item\textbf{Normalized Latency \citep{orca,vllmsosp}}: Total request execution time divided by number of decode tokens ($(T_s + T_p + T_d)/N_d$). Used for throughput comparison, with lower values being desirable at a given query rate.

\item\textbf{Capacity \citep{agrawal2024taming}}: Maximum sustainable request load (queries-per-second) while meeting defined latency SLOs. Higher capacity reduces serving costs.
\end{itemize}

\subsection{Challenges in Evaluating LLM Inference Serving Systems}
\label{sec:back:evalchallenges}

Evaluating LLM inference systems poses unique difficulties exceeding those in traditional ML serving scenarios. These stem from the inherent complexity of the inference process, diverse application needs, and the field's rapid evolution.

\vheading{Dual-Phase Nature of LLM Inference Requests} LLM inference workloads exhibit fundamentally different characteristics between their prefill and decode phases. The prefill phase is compute-bound \citep{sarathi2023}, processing hundreds to thousands of input tokens in parallel with quadratic computational complexity. In contrast, the decode phase is memory-bound, generating one token at a time with intensive memory access patterns due to KV-cache requirements. This duality creates complex performance dynamics with high inter-request variability in resource utilization. Furthermore, the prefill phase's quadratic time complexity with input length creates a fundamental tension between user expectations of consistent performance and what is computationally feasible. For instance, while users might expect similar response times for all inputs, a prompt of 32K tokens requires 4\myx more computation than a 16K token prompt. This variable cost structure makes it challenging to define meaningful performance targets.

\vspace{-0.5em}

\begin{table*}[h!]
\centering
\begin{tabularx}{0.95\textwidth}{l|ccc|ccc|cc}
 \toprule
 \multirow{2}{*}{\textbf{Trace}} & \multicolumn{3}{c|}{\textbf{\# Prefill tokens}} & \multicolumn{3}{c|}{\textbf{\# Decode tokens}} & \multicolumn{2}{c}{\textbf{P:D ratio}} \\
 & Median & IQR & P99 & Median & IQR & P99 & Median & IQR \\
 \midrule
 Azure Code 2024 & 1928 & 2393 & 7685 & 8 & 15 & 276 & 238 & 686 \\
 Azure Conv 2024 & 928 & 1811 & 6683 & 41 & 94 & 694 & 21 & 63 \\
 Mooncake & 6345 & 4243 & 61616 & 30 & 343 & 898 & 163.5 & 602 \\
 \bottomrule
\end{tabularx}
\caption{Details of public production traces from \cite{dynamollm, mooncake}.\vspace{-1em}}
\label{table:workloads}
\end{table*}

\vheading{Heterogeneous Application Requirements} 
Different LLM applications impose vastly different input characteristics and performance priorities, complicating the definition of "good" performance.  Our analysis of production traces in Table \ref{table:workloads} shows how request length distributions differ drastically across applications: Median prompt length differ by 2\myx between \textit{Code} and \textit{Conversation} applications in Azure LLM Inference 2024 dataset \citep{dynamollm}, while median output lengths differ by 5\myx. kimi.ai \citep{mooncake} receives median prompt lengths 6\myx longer than Azure Conversation. On the performance requirement side, Code completion tools might prioritize strict TTLT guarantees, tolerating some jitter, whereas conversational agents need consistent, low TBT for natural interaction flow, potentially sacrificing absolute throughput. Long-form content generation often prioritizes overall throughput (Capacity or low average TPOT) over immediate responsiveness (TTFT). These varying needs lead to different optimization goals and metric choices. Sensitivity to variance in latency also differs; advanced voice interfaces might accept lower token rates but have minimal tolerance for generation stalls that disrupt perceived fluidity.

\vheading{Fast-Changing Domain}
The rapid pace of LLM innovation further complicates evaluation. New architectures like Mixture of Experts (MoE)~\citep{shazeer2017outrageously} and Grouped Query Attention (GQA)~\citep{groupedqueryattention} exhibit different performance characteristics than traditional dense transformers. For instance, GQA models significantly reduce (e.g., 4-8x) the KV cache memory footprint, alleviating a major bottleneck that prompted innovations like PagedAttention~\citep{vllmsosp}. Such advances can render evaluation assumptions based on older architectures misleading or irrelevant. Furthermore, emerging applications continually introduce novel performance requirements and usage patterns, demanding that evaluation methodologies adapt continuously to remain meaningful and accurately assess performance across the evolving LLM landscape. This diversity underscores the need for robust, adaptable evaluation methodologies.

\begin{table*}[t!]
    \footnotesize
   \centering
   \begin{tabular}{l@{\hspace{4pt}}c@{\hspace{2pt}}c@{\hspace{1pt}}c@{\hspace{1pt}}c@{\hspace{1pt}}c@{\hspace{1pt}}c@{\hspace{1pt}}c@{\hspace{1pt}}c@{\hspace{1pt}}c@{\hspace{1pt}}c@{\hspace{1pt}}c@{\hspace{1pt}}c} 
   \toprule
   & \rotatebox{70}{Orca} & \rotatebox{70}{vLLM} & \rotatebox{70}{Sarathi} & \rotatebox{70}{DistServe} & \rotatebox{70}{dLoRA} & \rotatebox{70}{S-LoRA} & \rotatebox{70}{L2R} & \rotatebox{70}{MuxServe} & \rotatebox{70}{SpotServe} & \rotatebox{70}{LoongServe} & \rotatebox{70}{NanoFlow} & \rotatebox{70}{SplitWise} \\
   \noalign{\hrule height 0.5pt}
   \rowcolor{gray!15}
   \multicolumn{13}{c}{\textit{\textbf{Baseline Fairness}}}\\\hline
   Implementation Fairness & \greencheck & \greencheck & \greencheck & \redcross & \greencheck & \greencheck & \greencheck & \greencheck & \greencheck & \redcross & \greencheck & \greencheck \\
   Parameter Tuning & \greencheck & \redcross & \greencheck & \redcross & \greencheck & \greencheck & \greencheck & \greencheck & \greencheck & \redcross & \greencheck & \greencheck \\
   \noalign{\hrule height 0.5pt}
   \rowcolor{gray!15}
   \multicolumn{13}{c}{\textit{\textbf{Evaluation Setup}}}\\\hline
   Model Selection & \greencheck & \redcross & \greencheck & \greencheck & \greencheck & \redcross & \greencheck & \greencheck & \redcross & \greencheck & \greencheck & \greencheck \\
   Workload Diversity & \redcross & \redcross & \greencheck & \redcross & \redcross & \redcross & \redcross & \redcross & \redcross & \redcross & \greencheck & \greencheck \\
   Practical Latency & \greencheck & \greencheck & \redcross & \greencheck & \redcross & \redcross & \redcross & \redcross & \redcross & \redcross & \redcross & \greencheck \\
   \noalign{\hrule height 0.5pt}
   \rowcolor{gray!15}
   \multicolumn{13}{c}{\textit{\textbf{ Metrics Design}}}\\\hline
   Metric Selection & \greencheck & \greencheck & \greencheck & \greencheck & \redcross & \greencheck & \greencheck & \redcross & \greencheck & \greencheck & \greencheck & \greencheck \\
   Performance Distribution & \redcross & \redcross & \redcross & \greencheck & \redcross & \redcross & \greencheck & \redcross & \greencheck & \redcross & \greencheck & \greencheck\\
   No Obscuring Normalization & \redcross & \redcross & \greencheck & \redcross & \redcross & \greencheck & \redcross & \redcross & \greencheck & \redcross & \redcross & \redcross \\
   \bottomrule
   \end{tabular}
   \caption{Assessment of evaluation practices in recent LLM inference systems. Our analysis examines three key dimensions: baseline setup , evaluation methodology , and metric design. A checkmark (\greencheck) indicates the presence of a practice while a cross (\redcross) indicates its absence. \autoref{appendix:assesment} provides details of the assessment process. \vspace{-1em}}
   \label{tab:checklist-adherence}
\end{table*}

\section{Common Anti-Patterns in LLM Inference System Evaluations}
\label{sec:anti-patterns}

\subsection{Anti-Patterns in Baseline Fairness}
\label{sec:anti-patterns:fairness}

Comparing complex systems fairly requires meticulous attention to implementation details and configuration. Neglecting this leads to common anti-patterns where observed differences are misattributed.

\vheading{Anti-Pattern 1: Conflating Implementation and Algorithm}
LLM inference systems are intricate engineering artifacts maximizing accelerator utilization. Differences in underlying implementations (e.g., scheduling overheads, communication protocols, low-level optimizations) can significantly impact performance -- as shown by ~\cite{srivatsa2024scheduling} recently, the CPU overheads can contribute to $>$50\% latency. Failing to isolate algorithmic contributions from system engineering prowess is a frequent anti-pattern. It makes discerning the true benefit of a novel technique challenging, potentially misdirecting research efforts.

While systems like DistServe and LoongServe~\citep{distserve2024, 2024loongserve} compare against standard baselines (vLLM, DeepSpeed-MII), they often lack evaluations that rigorously separate algorithmic gains from implementation-specific advantages. In contrast, NanoFlow~\citep{nanoflow} provides exemplary transparency. While reporting a 1.91\myx throughput gain from their nano-batching technique, their careful ablation studies revealed that system-level optimizations accounted for most of this, with the core algorithmic idea contributing only 16\% when isolated. Such detailed breakdowns, crucial for understanding true innovation, remain rare. This anti-pattern allows implementation differences to mask and inaccurately amplify the impact of algorithmic changes.

\begin{checklistitem}
\textbf{Implementation Fairness:} Are baselines implemented comparably? If not, are microbenchmarks or ablations used to isolate algorithmic gains from system implementation differences?
\end{checklistitem}

\vheading{Anti-Pattern 2: Neglecting Parameter Tuning}
LLM serving systems expose rich configuration spaces (e.g., batch sizes, resource allocation ratios, scheduling policies) allowing optimization for diverse scenarios~\citep{vidur}. For instance, techniques like disaggregated prefill/decode~\citep{patel2023splitwise, distserve2024} require careful tuning of resource allocation for optimal performance. A common anti-pattern involves presenting results for a new system with extensive tuning while comparing against baselines using default or arbitrarily chosen parameters. Seemingly minor configuration differences can cause significant performance variations, invalidating the comparison.

For instance, comparisons against DeepSpeed-MII's chunked prefill in DistServe~\citep{distserve2024} and LoongServe~\citep{2024loongserve} did not report tuning the critical chunk size parameter for the baseline according to the workload, despite prior work showing its significant impact~\citep{sarathi2023}. Conversely, Orca~\citep{orca} demonstrates good practice by carefully evaluating its baseline (FasterTransformer \citep{fastertransformer}) across different batch sizes, ensuring a fairer comparison point. Failing to apply comparable tuning effort to baselines is a critical fairness anti-pattern.

\begin{checklistitem}
\textbf{Parameter Tuning:} Are all performance-critical configuration parameters documented for both the proposed system and baselines? Were baseline parameters tuned appropriately for the evaluation setup?
\end{checklistitem}

\subsection{Anti-Patterns in Evaluation Setup}
\label{sec:anti-patterns:setup}

The choice of models, workloads, and operating constraints significantly impacts results. Using setups unrepresentative of current technology or real-world use constitutes another set of anti-patterns.

\vheading{Anti-Pattern 3: Evaluating with Outdated or Irrelevant Models}
The LLM landscape evolves rapidly. Architectures like GQA~\citep{groupedqueryattention, multiqueryattention} and MoE~\citep{shazeer2017outrageously} have different performance bottlenecks compared to older, standard MHA (Multi-Head Attention) models. Notably, GQA models require 4-8\myx less KV cache memory. Evaluating techniques, especially memory optimizations, solely on MHA models~\citep{vllmsosp, 2024infinigen, cachegen} is an anti-pattern that risks optimizing for issues less relevant in state-of-the-art models. Techniques may show significant benefits on MHA models simply because the memory footprint is larger, an advantage that might shrink or disappear on GQA models. DuoAttention~\citep{duoattention} sets a good precedent by evaluating its KV cache compression on both MHA and GQA models, showing diminishing returns (2.55\myx vs. 1.67\myx reduction) on the more modern architecture, providing crucial context. Evaluations should reflect contemporary model architectures to ensure proposed techniques address relevant bottlenecks.

\begin{checklistitem}
\textbf{Model Selection:} Do the evaluated models reflect current state-of-the-art? Are key techniques evaluated across different model architectures where their impact might significantly vary?
\end{checklistitem}

\vheading{Anti-Pattern 4: Using Non-Representative Workloads}
Production LLMs serve diverse applications with distinct workload characteristics (e.g., code completion: short prompt/output, low latency need; chat: medium length, variable output; summarization: long prompt/output)~\citep{dynamollm, wang2023openchat}. An anti-pattern is evaluating systems predominantly on single-source datasets (like ShareGPT capturing short chat) or synthetic workloads~\citep{slora, spotserve, orca}, often with limited input lengths (e.g., average 1024 tokens~\citep{vllmsosp, muxserve}). This narrow focus risks missing critical performance behaviors that emerge only with longer sequences, different interaction patterns (e.g., multi-turn), or high request variance. As Vidur~\citep{vidur} demonstrated, optimizing for one workload type can lead to up to 2\myx lower performance on another. Evaluating across a diverse set of realistic workloads is essential to understand a system's true strengths, weaknesses, and generalizability.

\begin{checklistitem}
\textbf{Workload Diversity}: Is the system evaluated beyond a single dataset or synthetic traces? Are diverse applications (chat, code, summarization) with their characteristic distributions of prompt/output lengths and interaction models evaluated?
\end{checklistitem}

\vheading{Anti-Pattern 5: Ignoring Practical Latency Bounds}
While pursuing algorithmic advances is vital, presenting improvements without context within practical latency requirements is an anti-pattern. Users experience absolute latencies (TTFT, TBT), not just relative gains or normalized metrics. Recent latency-optimized systems~\cite{distserve2024, patel2023splitwise, agrawal2024taming} highlighted that throughput gains claimed by prior works~\cite{orca, vllmsosp} often failed to materialize in practice because they required operating points with unacceptably high tail latency.

Consider LoongServe~\cite{2024loongserve}, proposing optimizations for long contexts (up to ~500K tokens). While technically interesting, the evaluations report normalized latencies per input token in the hundreds of milliseconds. This translates to absolute TTFTs in the range of thousands to tens of thousands of seconds, far exceeding any practical threshold for interactive use cases. We find similar patterns where the normalized latency numbers reported in papers conceal several orders of magnitude growth in absolute latency~\citep{learningtorank,spotserve,dlora}. DistServe~\citep{distserve2024} exemplifies good practice by establishing clear, application-specific latency bounds (SLOs) and evaluating performance within those constraints. Improvements must be demonstrated within latency regimes relevant to the target application.

\begin{checklistitem}
\textbf{Practical Latency Bounds:} Are reported improvements achieved within latency bounds (e.g., TTFT, TBT SLOs) suitable for the target application(s)? Is the connection between reported metrics and absolute user-experienced latency clear?
\end{checklistitem}

\subsection{Anti-Patterns in Metric Design and Interpretation}
\label{sec:anti-patterns:metrics}

The choice and presentation of metrics profoundly influence conclusions. Several anti-patterns related to metric selection, aggregation, and normalization can obscure critical performance details.

\vheading{Anti-Pattern 6: Selecting Misleading or Opaque Metrics}
The request-level heterogeneity in LLM inference workloads makes it challenging to perform evaluations using traditional metrics. Each request can vary dramatically in its characteristics - from prompt length and output length to latency sensitivity and resource requirements \autoref{sec:back:evalchallenges}. This complexity has led to a natural fragmentation in how the community evaluates LLM inference performance, with different papers adopting various metrics. This inconsistency makes it difficult to track progress and compare systems fairly. More importantly, in some cases, the chosen metrics may not align well with actual user experience.

For instance, dLoRA \citep{dlora} reports average latency, which is calculated by dividing the sum of each request’s end-to-end latency by the total number of output tokens. While convenient, this metric does not correspond to actual latency experienced by any user. This approach can also mask significant tail latency issues that severely impact user experience, making it unsuitable for robust evaluation.

\begin{checklistitem}
\textbf{Metric Selection}: Are the chosen metrics broadly used and understood? For a novel metric, is its relationship to user-perceived performance explained and justified?
\end{checklistitem}

\begin{figure*}[t!]
\centering
\begin{subfigure}[b]{0.25\textwidth}
    \centering
    \includegraphics[width=\textwidth]{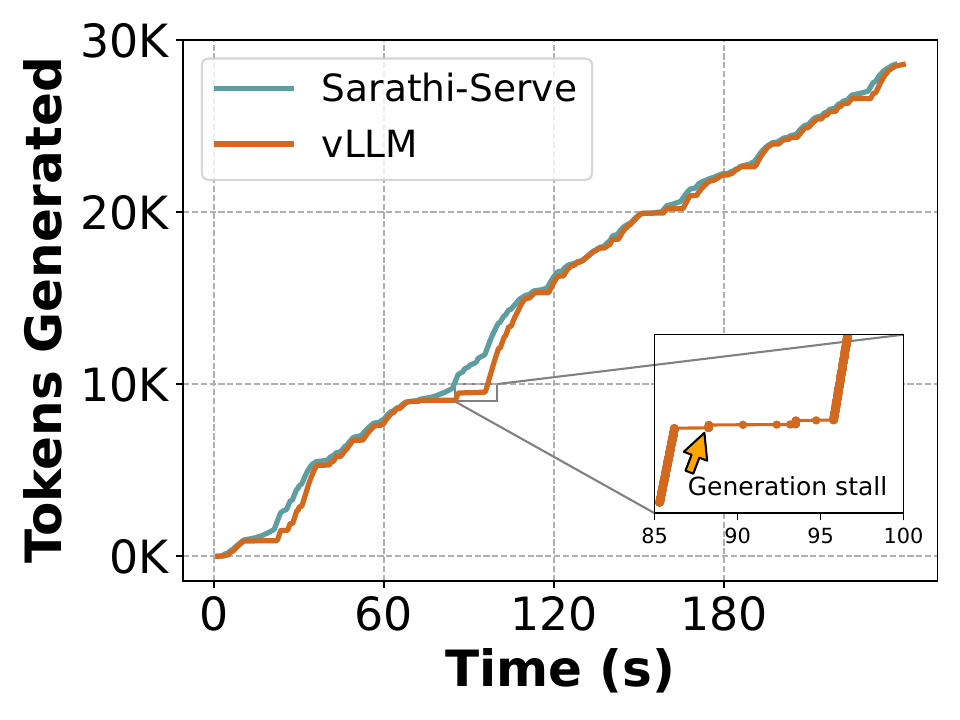}
    \caption{\small Generation stalls in output tokens generation.}
    \label{fig:mot:tbt:timeline}
\end{subfigure}
\quad
\begin{subfigure}[b]{0.28\textwidth}
    \centering
    \includegraphics[width=\textwidth]{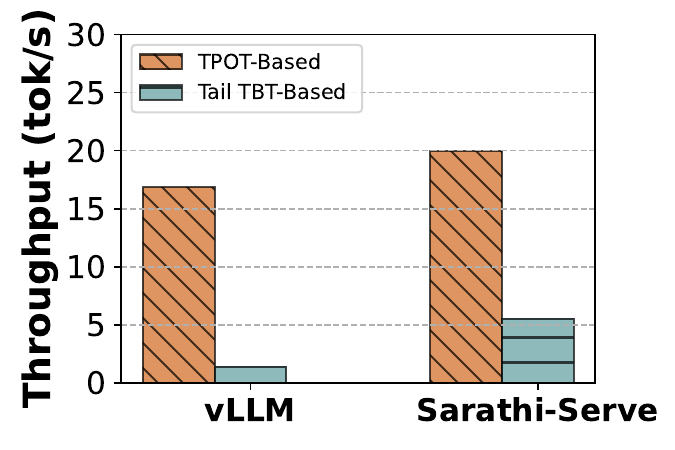}
    \caption{\small Token throughput derived using various latency metrics.}
    \label{fig:mot:tbt:tput}
\end{subfigure}
\quad
\begin{subfigure}[b]{0.33\textwidth}
    \centering
    \includegraphics[width=\textwidth]{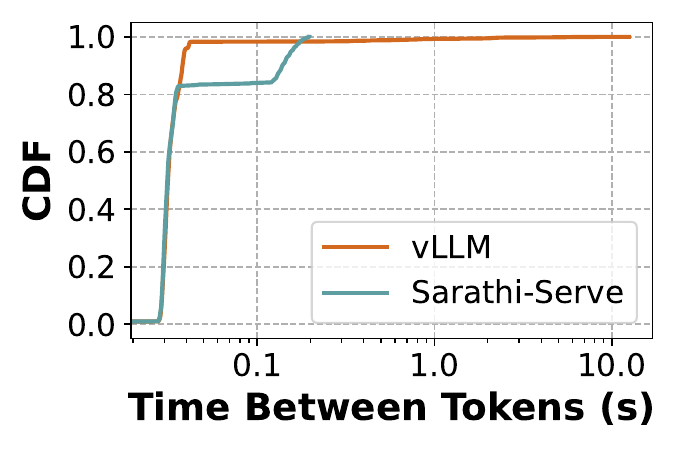}
    \caption{\small Decode token latency distribution.}
    \label{fig:mot:tbt:cdf}
\end{subfigure}
\caption{Limitations of conventional metrics illustrated with vLLM and Sarathi-Serve.
(a)~Actual token generation may include significant stalls.
(b)~Averaging metrics like TPOT hide these stalls, potentially overestimating effective throughput.
(c)~The full TBT distribution reveals nuances (e.g., P85 vs. P99 trade-offs) missed by tail-only analysis.}
\label{fig:mot:tbt}
\end{figure*}

\vheading{Anti-Pattern 7: Reporting Only Summary Statistics (Ignoring Distribution)}
Focusing solely on summary statistics (like median latency) while ignoring the full performance distribution is a prevalent anti-pattern. Early systems like vLLM~\citep{vllmsosp} and Orca~\citep{orca} primarily reported median latency. Subsequent work~\citep{distserve2024, agrawal2024taming} revealed these systems suffered from high tail latency, hindering production deployment. However, reacting by focusing \textit{only} on tail latency (e.g., P99) is also incomplete. As shown in \autoref{fig:mot:tbt:cdf}, comparing Sarathi-Serve and vLLM on the \arxivsum dataset reveals a trade-off: Sarathi-Serve achieves better P99 latency but exhibits higher latency around the P90 percentile. Such crucial trade-offs, vital for understanding system behavior under load and selecting appropriate systems for specific SLOs, are only visible when examining the full performance distribution (or at least multiple key percentiles like P50, P90, P99).

\begin{checklistitem}
\textbf{Performance Distribution}: Does the evaluation present the full latency distribution (e.g., CDFs) or at least key percentiles (P50, P90, P99), rather than just a single summary statistic like average or median? Does the analysis illuminate performance trade-offs across different points in the distribution?
\end{checklistitem}
\vspace{0.5em}

\vheading{Anti-Pattern 8: Obscuring Performance with Normalization}
Normalization is standard practice, but its application in LLM inference can inadvertently create an anti-pattern by obscuring user-facing issues. Metrics like normalized latency ($(T_s + T_p + T_d)/N_d$)~\citep{orca, vllmsosp, spotserve, 2024loongserve} and TPOT ($T_d/N_d$)~\citep{distserve2024, patel2023splitwise, learningtorank} are common. However, they can mask the impact of fixed or near-fixed overheads, especially scheduling delay ($T_s$).

Consider an example: two requests generating 10 and 1000 tokens respectively, both experience a 10s scheduling delay, with 20ms per-token generation. Normalized latency is dramatically different (1.02 s/token vs. 0.03 s/token), falsely suggesting a vast difference in user experience. In reality, both users suffer the same unacceptable 10s initial wait. Normalization dilutes the fixed delay based on the (variable) number of output tokens. \autoref{fig:mot:tbt} empirically confirms this: evaluating Sarathi-Serve and vLLM on \arxivsum shows vLLM having $>$25s scheduling delays for 60\% of requests, yet this critical difference is nearly invisible in the normalized latency plot.

Similarly, TPOT (average decode time) masks temporal inconsistencies. As \autoref{fig:mot:tbt:timeline} illustrates, two systems can have identical TPOT, but one might generate tokens smoothly while the other produces them in bursts with long stalls in between. These stalls, hidden by averaging, severely degrade the user experience in interactive applications. While normalization has its utility, relying on it exclusively without examining raw, absolute performance characteristics (like $T_s$ and TBT distribution) is a significant anti-pattern.

\vspace{1em}

\begin{checklistitem}
\textbf{No Obscuring Metric Normalization}: When using normalized metrics (Normalized Latency, TPOT), are raw performance characteristics also examined? Are critical temporal aspects like scheduling delays ($T_s$) and token generation stalls (TBT variance) analyzed independently? Does the evaluation consider how normalization might obscure user-facing performance issues evident in raw measurements?
\end{checklistitem}

\begin{figure}[t!]
\centering
\begin{subfigure}[b]{0.31\linewidth}
    \centering
    \includegraphics[width=\linewidth]{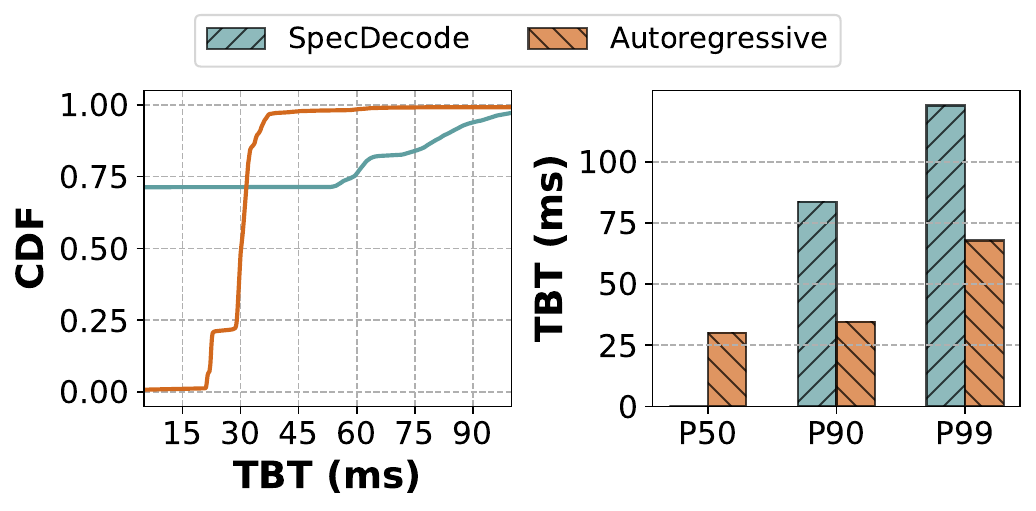}
    \caption{TBT latency characteristics showing bursty token generation in speculative decoding. CDF (left) reveals 75\% tokens with near-zero TBT due to batch acceptance.} %
    \label{fig:eval:specdec:tbt}
\end{subfigure}
\quad
\begin{subfigure}[b]{0.31\linewidth}
    \centering
    \includegraphics[width=\linewidth]{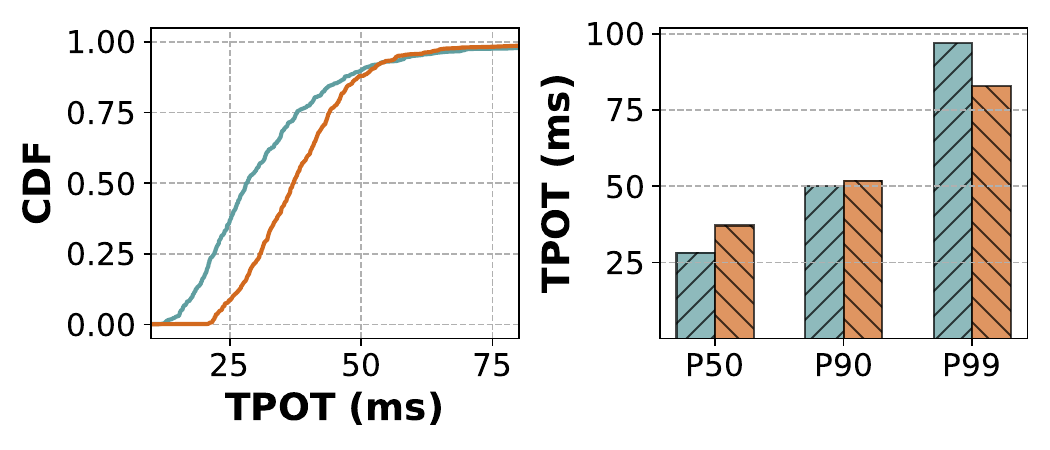}
    \caption{TPOT analysis showing the aggregate latency trends: lower median costs but higher tail latency. \vspace{3em}}
    \label{fig:eval:specdec:tpot}
\end{subfigure}
\quad
\begin{subfigure}[b]{0.31\linewidth}
    \centering
    \includegraphics[width=\linewidth]{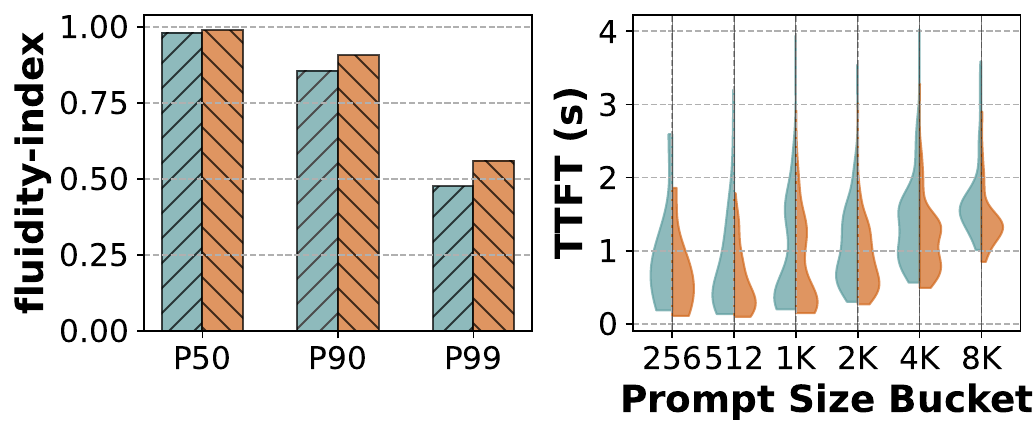}
    \caption{Fluidity-index values at different percentiles (left) and TTFT distributions across prompt sizes (right), showing higher TTFT for speculative decoding due to additional prefill overhead for drafter. }
    \label{fig:eval:specdec:fittft}
\end{subfigure}
\caption{Speculative decoding vs. autoregressive performance (0.25 QPS, ShareGPT4, Llama-3 70B, 4xH100). The figure highlights impact common evaluation anti-patterns: (a)~misleading TBT distributions, (b)~conflicting TPOT trends (median vs. tail), and (c, right)~obscured TTFT impact due to draft overhead. These demonstrate how isolated conventional metrics misrepresent speculative decoding's trade-offs; incorporating consistency metrics like Fluidity Index (c, left), is required for accurate assessment.}
\label{fig:eval:specdec}
\end{figure}

\section{Case Study: Evaluating Speculative Decoding Systems}
\label{sec:case-study}

To illustrate the practical consequences of evaluation anti-patterns and the benefits of a rigorous methodology, we examine speculative decoding -- a popular latency optimization technique, that exhibits non-uniform token-generations latencies, which makes it challenging to evaluate using standard metrics. Speculative decoding~\citep{pmlr-v202-leviathan23a, chen2023accelerating} employs a smaller "draft" model to predict multiple future tokens, which are then verified in parallel by the larger primary model. This results in a spread of decode latencies: successful verifications yield bursts of tokens arriving almost instantly, while verification failures lead to delays as only a single token is produced after incurring costs for both draft generation and verification.

Evaluating this technique using standard metrics often exemplifies the anti-patterns discussed previously (\autoref{sec:anti-patterns}), leading to incomplete or misleading conclusions. Our analysis, illustrated in part by \autoref{fig:eval:specdec}, reveals these limitations: 

\textbf{TBT becomes misleading:} The TBT distribution becomes highly bimodal. As seen in \autoref{fig:eval:specdec:tbt}, a large fraction (e.g., ~75\%) of inter-token intervals register near-zero latency due to batch acceptance. Reporting only a median TBT would capture only this best-case scenario, ignoring the significant delays during verification failures---a clear example of \textbf{Anti-Pattern 7} (Ignoring Distribution).

\textbf{TPOT shows conflicting trends:} While speculative decoding might appear favorable based on median TPOT (e.g., 1.3\myx lower in our experiments, \autoref{fig:eval:specdec:tpot}), this advantage can reverse at the tail (e.g., 1.16\myx higher P99 TPOT). Relying on just one summary statistic provides an incomplete picture, again falling into \textbf{Anti-Pattern 7}, while the averaging inherent in TPOT also masks the underlying variability, touching upon \textbf{Anti-Pattern 8}.

\textbf{TTFT is negatively impacted:} Speculative decoding consistently incurs higher TTFT (\autoref{fig:eval:specdec:fittft}) due to the additional prefill overhead required for the draft model. This critical factor for interactive responsiveness can be obscured if evaluations focus solely on throughput-oriented metrics like average TPOT, demonstrating aspects of \textbf{Anti-Pattern 5} (Ignoring Practical Latency Bounds) and \textbf{Anti-Pattern 8} (Obscuring Performance with Normalization, as throughput metrics average over the initial delay).

This case study vividly demonstrates how focusing on individual, conventional metrics can lead to contradictory conclusions, characteristic of the evaluation anti-patterns identified in \autoref{sec:anti-patterns}. Which metric should we trust? A more holistic evaluation, avoiding these anti-patterns, is necessary. For instance, incorporating metrics designed to capture generation consistency despite burstiness, such as the deadline-based Fluidity Index proposed by ~\cite{agrawal2024taming}, reveals a more nuanced picture. While speculative decoding often improves the \textit{average} token generation rate (reflected in lower median TPOT), such deadline-based metrics expose the introduced \textit{variability} and the increased likelihood of user-visible stalls during verification failures (contributing to the higher P99 TPOT). This highlights a critical trade-off between average speed and generation consistency, a trade-off particularly important for interactive applications sensitive to stalls. The overall performance profile also depends heavily on factors like the draft model's accuracy and the typical response length, where benefits increase for longer generations as the initial prefill overhead is amortized.

\vspace{-1em}
\section{Conclusion}
\label{sec:conclusion}

The evolution of LLM inference systems has outpaced our ability to evaluate them effectively, leading to critical \textit{anti-patterns} between reported performance and real-world user experience. Through systematic analysis of recent systems, we have identified recurring  in evaluation methodology spanning baseline fairness, experimental design, and metric selection—challenges uniquely problematic for LLM systems due to their dual-phase inference nature, heterogeneous workload characteristics, and latency requirements. Our comprehensive evaluation checklist provides researchers with critical guidance for meaningful performance assessment, as demonstrated through our case study of speculative decoding where proper methodology revealed nuanced performance trade-offs that traditional metrics often obscure. As the field continues its rapid evolution, these methodological principles offer a foundation for more meaningful comparisons, reproducible results, and serving systems that genuinely reflect real-world performance requirements, ultimately accelerating progress by ensuring that reported improvements translate to genuine user experience enhancements.
\\\\

\bibliography{main}
\bibliographystyle{colm2025_conference}

\clearpage

\appendix
\section*{Appendix A: Detailed System Assessment Methodology}
\label{appendix:assesment}

We present a detailed explanation of our assessment methodology for each system in \Cref{tab:checklist-adherence}. Our goal is to provide transparency about how we arrived at each assessment.

\subsection*{A.1 Assessment Process}

Our assessment is based on publicly available materials from each system, examining three key aspects shown in \Cref{tab:checklist-adherence}: baseline fairness, evaluation setup, and metric design.

\subsection*{A.1 System-by-System Analysis}

\textbf{\textit{{Orca \citep{orca}}}}
\hrule

\textbf{Implementation Fairness (\greencheck)}  The paper explicitly compares their engine with FasterTransformer (FT) \cite{fastertransformer} using microbenchmarks, demonstrating 30-40\% lower latency for 175B model sizes from engine improvements. Implementation-specific gains from algorithmic benefits through controlled experiments are clearly separated.

\textbf{Parameter Tuning (\greencheck)} The evaluation demonstrates results with different batch sizes in FasterTransformer. This thorough exploration of baseline configurations supports fair comparison.

\textbf{Model Selection (\greencheck)} The evaluation spans four model sizes (13B, 101B, 175B, 341B) and considers both tensor and pipeline parallelism strategies. This range represented state-of-the-art architectures at publication time.

\textbf{Workload Diversity (\redcross)} The evaluation uses only short requests with prefill lengths uniformly distributed between 32-512 tokens and decode lengths between 1-128 tokens. This limited range doesn't capture the longer request sizes common in production deployments.

\textbf{Practical Latency (\greencheck)} Results focus on the practical operating regime below 100ms/token, using logarithmic scales to clearly show differences in critical regions. While Figure 11 includes some high latency results (1000ms, 10000ms), these are appropriately contextualized.

\textbf{Metric Selection (\greencheck)} The paper uses median normalized end-to-end latency. While normalization obfuscates some information, the metric is well intuitive and commonly used.

\textbf{Performance Distribution (\redcross)} The analysis reports only median values without examining tail latency behavior, missing important reliability characteristics.

\textbf{No Obscuring Normalization (\redcross)} The evaluation normalizes entire end-to-end latency by output length, potentially masking scheduling delays that impact user experience.

\vspace{1em}
\textbf{\textit{{vLLM \citep{vllmsosp}}}}
\hrule

\textbf{Implementation Fairness (\greencheck)} The authors implement FT, Orca, and vLLM on the same codebase, enabling direct comparison of algorithmic benefits. This shared implementation foundation strengthens comparative results.

\textbf{Parameter Tuning (\redcross)} The evaluation uses a fixed batch size of 8 for all experiments without exploring optimal configurations for baselines. This limitation was only discovered through artifact examination, not documented in the paper.

\textbf{Model Selection (\redcross)} Despite GQA models being available, the evaluation uses only MHA models across sizes (13B, 66B, 175B). MHA models have higher memory KV cache footprint compared to GQA models, which provides additional advatange to PagedAttention.

\textbf{Workload Diversity (\redcross)} The evaluation focuses on short sequences, with ShareGPT (mean prefill=161, decode=331) and Alpaca (mean prefill=19, decode=58) datasets.

\textbf{Practical Latency (\greencheck)} Most reported measurements maintain sub-100ms latency, with clear visualization of performance in practical operating regions.

\textbf{Metric Selection (\greencheck)}
The paper uses median normalized end-to-end latency, which was standard practice at the time and suitable for throughput-oriented evaluation.

\textbf{Performance Distribution (\redcross)}
Only reports median values, missing crucial tail latency analysis that would reveal system stability under load.

\textbf{No Obscuring Normalization (\redcross)}
Uses normalized latency metric which hides large scheduling delays and generation stalls.

\vspace{1em}
\textbf{\textit{{Sarathi-Serve \citep{agrawal2024taming}}}}
\hrule

\textbf{Implementation Fairness (\greencheck)} All baseline schedulers (Orca, vLLM, Sarathi) share a common codebase, facilitating fair technical comparison.

\textbf{Parameter Tuning (\greencheck)} The evaluation tests multiple batch sizes in vLLM and documents configuration choices for preventing memory issues.

\textbf{Model Selection (\greencheck)} The analysis covers diverse model sizes (7B, 34B, 70B, 180B) including both MHA and GQA architectures, representing the state-of-the-art model architectures.

\textbf{Workload Diversity (\greencheck)} The evaluation is performed over ShareGPT4 and \arxivsum traces, both of which are distinct in nature and exhibit large request length variations.

\textbf{Practical Latency (\redcross)} The evaluation employs loose TBT P99 bounds of 5s in the ``relaxed" setting that exceed practical usability thresholds for interactive applications.

\textbf{Metric Selection (\greencheck)}
Uses Time Between Tokens (TBT) as the primary metric, which directly corresponds to user-perceived generation fluidity. This choice appropriately reflects real-world requirements for interactive applications.

\textbf{Performance Distribution (\redcross)}
While the paper extensively analyzes P99 TBT, it lacks discussion of median TBT behavior.

\textbf{No Obscuring Normalization (\greencheck)}
Avoids normalization pitfalls by directly measuring raw latencies like TBT and TTFT. This approach better preserves visibility into actual user experience.

\vspace{1em}
\textbf{\textit{{DistServe \citep{distserve2024}}}}
\hrule

\textbf{Implementation Fairness (\redcross)} The evaluation uses three different codebases (vLLM, DistServe, DeepSpeed-MII) without microbenchmarks to isolate implementation differences from algorithmic benefits.

\textbf{Parameter Tuning (\redcross)} Critical configuration parameters like chunk size for DeepSpeed-MII and batch sizes for vLLM are not documented, making it difficult to verify fair comparison.

\textbf{Model Selection (\greencheck)} The evaluation uses a good spread of model sizes with OPT-13B, 66B, and 175B, representing different computational requirements.

\textbf{Workload Diversity (\redcross)} Maximum prefill length is limited to 2K tokens (LongBench), with even shorter decode lengths (maximum average 200 tokens), missing important long-context scenarios.

\textbf{Practical Latency (\greencheck)} Both TTFT and TPOT targets are reasonable and workload-dependent, reflecting real-world requirements.

\textbf{Metric Selection (\greencheck)}
Uses percentage of requests meeting SLO, considering both TTFT and TPOT. However, notably omits analysis of Time to Second Token, which would be significantly impacted by their disaggregation approach. This metric deserves attention given the architectural choice of separating prefill and decode phases.

\textbf{Performance Distribution (\greencheck)}
Shows both 90\% and 99\% SLO attainment, providing a more complete view of system reliability. The evaluation also demonstrates how SLO attainment varies with increasing request rate, offering insights into system behavior under load.

\textbf{No Obscuring Normalization (\redcross)}
Relies on TPOT (Time Per Output Token), which can mask generation stalls in their disaggregated architecture. This is particularly concerning for effects like second token generation delays, which can be hidden by averaging across all output tokens despite significantly impacting user experience.

\vspace{1em}
\textbf{\textit{{dLoRA \citep{dlora}}}}
\hrule

\textbf{Implementation Fairness (\greencheck)}
Implements baselines (vLLM with uniform allocation and PEFT) in the same codebase with key optimizations like Selective Batching and PagedAttention.

\textbf{Parameter Tuning (\greencheck)}
While specific tuning details are limited in the paper, the evaluation demonstrates consideration of key configuration parameters for each system.

\textbf{Model Selection (\greencheck)}
Evaluates three sizes of Llama-2 (7B, 13B, 70B) with appropriate tensor parallelism configurations. The choice of Llama models is well-justified given their popularity in fine-tuning scenarios.

\textbf{Workload Diversity (\redcross)}
Uses only ShareGPT trace with relatively short chat queries, missing evaluation on longer context workloads or different application patterns.

\textbf{Practical Latency (\redcross)}
Uses average normalized request  latency with an SLO of 500ms per token, which is impractically high.

\textbf{Metric Selection (\redcross)}
Calculates average latency as (sum of latencies of all requests / sum of output lengths of all requests), which provides an overly broad averaging that masks important performance characteristics.

\textbf{Performance Distribution (\redcross)}
Reports only average latency values, missing information about tail behavior.

\textbf{No Obscuring Normalization (\redcross)}
The broad averaging approach in their latency calculations obscures both scheduling delays and runtime variations.

\vspace{1em}
\textbf{\textit{{S-LoRA \citep{slora}}}}
\hrule

\textbf{Implementation Fairness (\greencheck)}
Most of the evaluations are focused on various baselines implemented with the S-LoRA framework.

\textbf{Parameter Tuning (\greencheck)}
Configuration tuning considerations are adequately addressed within the scope of their evaluation.

\textbf{Model Selection (\redcross)}
While using multiple Llama model sizes (7B, 13B, 30B, 70B) with different adapter ranks, they specifically disable GQA for the 70B model against official architecture specifications. This choice artificially increases memory pressure, potentially favoring their solution.

\textbf{Workload Diversity (\redcross)}
Uses LMSys trace (average prefill 85 tokens, decode 165 tokens) and synthetic trace with uniform(8, 512) distribution. Both represent relatively short sequences, missing important long-context scenarios.

\textbf{Practical Latency (\redcross)}
Uses 6s TTFT as SLO for small prefills, which exceeds practical usability thresholds.

\textbf{Metric Selection (\greencheck)}
Uses appropriate metrics including request latency and TTFT.

\textbf{Performance Distribution (\redcross)}
Reports only average values for request latency and TTFT, missing tail latency analysis.

\textbf{No Obscuring Normalization (\greencheck)}
Avoids normalization pitfalls by measuring raw latencies.

\vspace{1em}
\textbf{\textit{{Learning to Rank \citep{learningtorank}}}}
\hrule

\textbf{Implementation Fairness (\greencheck)}
All baselines implemented on the same codebase (vLLM), enabling fair comparison.

\textbf{Parameter Tuning (\greencheck)}
Documents configuration choices clearly, including token generation parameters for Perception baseline and bucket configurations for Classification baseline.

\textbf{Model Selection (\greencheck)}
Uses Llama3 8B (TP1) and 70B (TP8), representing current open-weight LLM standards.

\textbf{Workload Diversity (\redcross)}
Limited to short chat traces (LMSys1M and ShareGPT), missing evaluation on other workload types.

\textbf{Practical Latency (\redcross)}
Shows normalized latency in several seconds, far exceeding practical bounds.

\textbf{Metric Selection (\greencheck)}
Uses normalized latency, a commonly accepted metric.

\textbf{Performance Distribution (\greencheck)}
Shows both average and P90 normalized latency values.

\textbf{No Obscuring Normalization (\redcross)}
Normalized latency masks potentially large scheduling delays at the operating points evaluated in the paper.

\vspace{1em}
\textbf{\textit{{MuxServe \citep{muxserve}}}}
\hrule

\textbf{Implementation Fairness (\greencheck)}
Both baselines and proposed system are built on vLLM.

\textbf{Parameter Tuning (\greencheck)}
Deployment of AlpaServe is optimized with the same placement algorithm, ensuring fair comparison.

\textbf{Model Selection (\greencheck)}
Evaluates Llama models from four different size categories, providing good coverage of computational requirements.

\textbf{Workload Diversity (\redcross)}
Limited to ShareGPT and LMSys-Chat traces with small prefill and decode lengths, missing evaluation on longer sequences.

\textbf{Practical Latency (\redcross)}
Latency SLO combinations that achieve reasonable attainment rate are in non-interactive ranges.

\textbf{Metric Selection (\redcross)}
Main paper doesn't specify which latency metric is used -- ambiguous term ``latency" is used across the paper without specifying which latency.

\textbf{Performance Distribution (\redcross)}
Main section lacks discussion of latency metrics and percentiles, though appendix includes some P99 analysis.

\textbf{No Obscuring Normalization (\redcross)}
The latency metric used in the main body of paper is unclear. The appendix uses TPOT which can mask generation stalls.

\vspace{1em}
\textbf{\textit{{SpotServe \citep{spotserve}}}}
\hrule

\textbf{Implementation Fairness (\greencheck)}
Implements the proposed system and the two baselines on the same codebase for fairness.

\textbf{Parameter Tuning (\greencheck)} In our judgment, the baselines don't need parameter tuning. 

\textbf{Model Selection (\redcross)}
We appreciate evaluation on three models with different sizes: OPT-6.7B, GPT-20B and Llama-30B. These are running on severely memory constrained 4x T4 GPU nodes with 64GB memory capacity per node. In these settings, running Llama-30B is not a suitable, as demonstrated by the stable rate of 0.2 QPS used for this model. A tiny model like Phi-2-2.7B or a GQA model of size comparable to 30B would have been more practical.

\textbf{Workload Diversity (\redcross)}
The evaluation uses a single request length of 512 prefill tokens and 128 decode tokens. Using a trace with high skew might have made the evaluation more complicated to analyze as the number of available nodes keeps changing, unlike other inference systems. However, other request sizes should also have been evaluated.

\textbf{Practical Latency (\redcross)}
For Llama-30B, the E2E request latency for the best system is approximately 50s on several traces. This amounts to 390ms per output token, which is not desirable in online inference.  

\textbf{Metric Selection (\greencheck)}
Uses E2E request latency, which is well understood. Since the request length is small (512 prefill tokens and 128 decode tokens), using E2E and not mentioning TTFT and TPOT is fine.

\textbf{Performance Distribution (\greencheck)}
The paper does a great job here by showing average and several percentiles from 90th to 99th percentile.

\textbf{No Obscuring Normalization (\greencheck)}
Normalized metrics are not used.

\vspace{1em}
\textbf{\textit{{LoongServe \citep{2024loongserve}}}}
\hrule

\textbf{Implementation Fairness (\redcross)}
Uses three different codebases (vLLM, LightLLM, DeepSpeed-MII) without microbenchmarks to isolate implementation differences.

\textbf{Parameter Tuning (\redcross)}
Chunk size for DeepSpeed-MII is not specified. LightLLM chunk size uses formula from Sarathi \cite{sarathi2023} despite different operating context -- the original formulation was designed for offline inference scenarios for throughput maximization.

\textbf{Model Selection (\greencheck)}
While only testing one model (Llama-2 7B), it was a reasonable choice given its prominence at publication time. However, inclusion of more model sizes could have helped with more rounded evaluations.

\textbf{Workload Diversity (\redcross)}
Evaluations includes multiple datasets with varying prefill lengths. However, all long-context datasets have extremely small decode lengths (P90 decode tokens $<50$), limiting evaluation of key techniques like multi-master distributed decoding.

\textbf{Practical Latency (\redcross)}
The normalized input token latency in main evaluation corresponds to thousands of seconds of TTFT, far beyond practical bounds.

\textbf{Metric Selection (\greencheck)}
Uses normalized latency, which while not ideal, is well understood.

\textbf{Performance Distribution (\redcross)}
Reports only average numbers without tail analysis.

\textbf{No Obscuring Normalization (\redcross)}
All latency numbers are normalized, potentially masking important performance characteristics.

\vspace{1em}
\textbf{\textit{{NanoFlow \citep{nanoflow}}}}
\hrule

\textbf{Implementation Fairness (\greencheck)} While using multiple codebases (vLLM, TensorRT-LLM, DeepSpeed-MII, NanoFlow), the authors provide careful ablation studies isolating implementation differences from core algorithmic benefits.

\textbf{Parameter Tuning (\greencheck)} The evaluation uses optimal parameters for each baseline: best-performing max\_seq\_len for vLLM and max-ragged-batch-size for DeepSpeed-MII.

\textbf{Model Selection (\greencheck)} Tests six different models from 8B to 72B, including an MoE model, representing current architectural diversity.

\textbf{Workload Diversity (\greencheck)} Uses multiple  conversation traces, along with ablations with synthetic traces of varying prefill to decode ratios. Evaluation could have benefited from some longer context datasets.

\textbf{Practical Latency (\redcross)} Targets 200ms normalized latency, which exceeds practical bounds for interactive applications.

\textbf{Metric Selection (\greencheck)}
Uses normalized latency, which while not ideal, is a commonly understood metric in the community.

\textbf{Performance Distribution (\greencheck)}
While mostly reporting average normalized latency, they do provide a full CDF of normalized latency across three traces for a specific 90\% capacity setting, offering some insight into performance distribution.

\textbf{No Obscuring Normalization (\redcross)}
The use of normalized latency by output length masks potentially large scheduling delays.

\vspace{1em}
\textbf{\textit{Splitwise \citep{patel2023splitwise} }}
\hrule

\textbf{Implementation Fairness (\greencheck)}
All baselines are run with same backend, enabling fair comparison.

\textbf{Parameter Tuning (\greencheck)}
Enhances vLLM with mixed batching support to reduce generation stalls, demonstrating attention to baseline optimization.

\textbf{Model Selection (\greencheck)}
Uses Llama2-70B (highly popular) and BLOOM-176B. While BLOOM is somewhat dated, the selection represents different architectural choices.

\textbf{Workload Diversity (\greencheck)}
Uses real traces from Azure covering both conversation and coding workloads, with prompts up to 8K tokens and reasonable decode lengths (P99 500 tokens for conversation).

\textbf{Practical Latency (\greencheck)}
P50 limits are 1.25\myx single request execution time, with P99 up to 5\myx. Authors acknowledge slightly loose TTFT SLOs.

\textbf{Metric Selection (\greencheck)}
Uses TTFT, TBT and end-to-end latency, providing comprehensive performance view.

\textbf{Performance Distribution (\greencheck)}
Shows P50, P90 and P99 for all three latency metrics.

\textbf{No Obscuring Normalization (\redcross)} Uses TPOT for measuring decode latency (note that, this is referred as TBT in the Splitwise paper), which can obscure latency variations, specially the time to second token, which is affected due to disaggregation.

\subsection*{A.2 Limitations}

Our evaluation process has inherent limitations. We rely primarily on published materials, which may not capture full system capabilities. The rapid evolution of LLM serving means that there often no established standards when new research articles are originally proposed. Additionally, implementation details crucial to fair comparison may be incompletely documented.

\end{document}